\documentclass[10pt,twocolumn,letterpaper]{article}

\usepackage{cvpr}              
\usepackage{graphicx}
\usepackage{amsmath}
\usepackage{amssymb}
\usepackage{booktabs}
\usepackage{appendix}
\usepackage{multicol}
\usepackage{amsfonts}
\usepackage{bibentry}
\usepackage{multirow}
\usepackage{amssymb}
\usepackage{pifont}
\usepackage{bbding}
\usepackage[accsupp]{axessibility}
\def\semichecked{\checkmark\!\!\!\raisebox{0.4 em}{\tiny$\smallsetminus$}}

\usepackage[pagebackref,breaklinks,colorlinks]{hyperref}
\newcommand{\tablestyle}[2]{\setlength{\tabcolsep}{#1}\renewcommand{\arraystretch}{#2}\centering\footnotesize}

\usepackage[capitalize]{cleveref}
\crefname{section}{Sec.}{Secs.}
\Crefname{section}{Section}{Sections}
\Crefname{table}{Table}{Tables}
\crefname{table}{Tab.}{Tabs.}

\def\cvprPaperID{3323} 
\def\confName{CVPR}
\def\confYear{2022}
\usepackage[english]{babel}

\begin{document}

\title{HL-Net: Heterophily Learning Network for Scene Graph Generation}

\author{Xin Lin$^1$\thanks{Work done during first author’s internship at JD Explore Academy} \quad Changxing Ding$^{1,2}$\thanks{Corresponding author.} \quad Yibing Zhan$^{3}$ \quad Zijian Li$^1$ \quad Dacheng Tao$^{3,4}$\\
$^1$ South China University of Technology \quad
$^2$ Pazhou Lab, Guangzhou \quad
$^3$ JD Explore Academy\quad\\
$^4$ The University of Sydney \\
{{\tt\small eelinxin@mail.scut.edu.cn,\quad
chxding@scut.edu.cn,\quad eezijianli@mail.scut.edu.cn,}} \\
{{\tt\small zhanyibing@jd.com,\quad
dacheng.tao@gmail.com}}}
\maketitle

\begin{abstract}
Scene graph generation (SGG) aims to detect objects and predict their pairwise relationships within an image. Current SGG methods typically utilize graph neural networks (GNNs) to acquire context information between objects/relationships. Despite their effectiveness, however, current SGG methods only assume scene graph homophily while ignoring heterophily. Accordingly, in this paper, we propose a novel Heterophily Learning Network (HL-Net) to comprehensively explore the homophily and heterophily between objects/relationships in scene graphs. More specifically, HL-Net comprises the following 1) an adaptive reweighting transformer module, which adaptively integrates the information from different layers to exploit both the heterophily and homophily in objects; 2) a relationship feature propagation module that efficiently explores the connections between relationships by considering heterophily in order to refine the relationship representation; 3) a heterophily-aware message-passing scheme to further distinguish the heterophily and homophily between objects/relationships, thereby facilitating improved message passing in graphs. We conducted extensive experiments on two public datasets: Visual Genome (VG) and Open Images (OI). The experimental results demonstrate the superiority of our proposed HL-Net over existing state-of-the-art approaches. In more detail, HL-Net outperforms the second-best competitors by 2.1$\%$ on the VG dataset for scene graph classification and 1.2$\%$ on the IO dataset for the final score. Code is available at \href{https://github.com/siml3/HL-Net}{
https://github.com/siml3/HL-Net}. 
\end{abstract}

\section{Introduction}\label{introduction}

\begin{figure}[tbp]
    \centering
    \includegraphics[width=1.\linewidth]{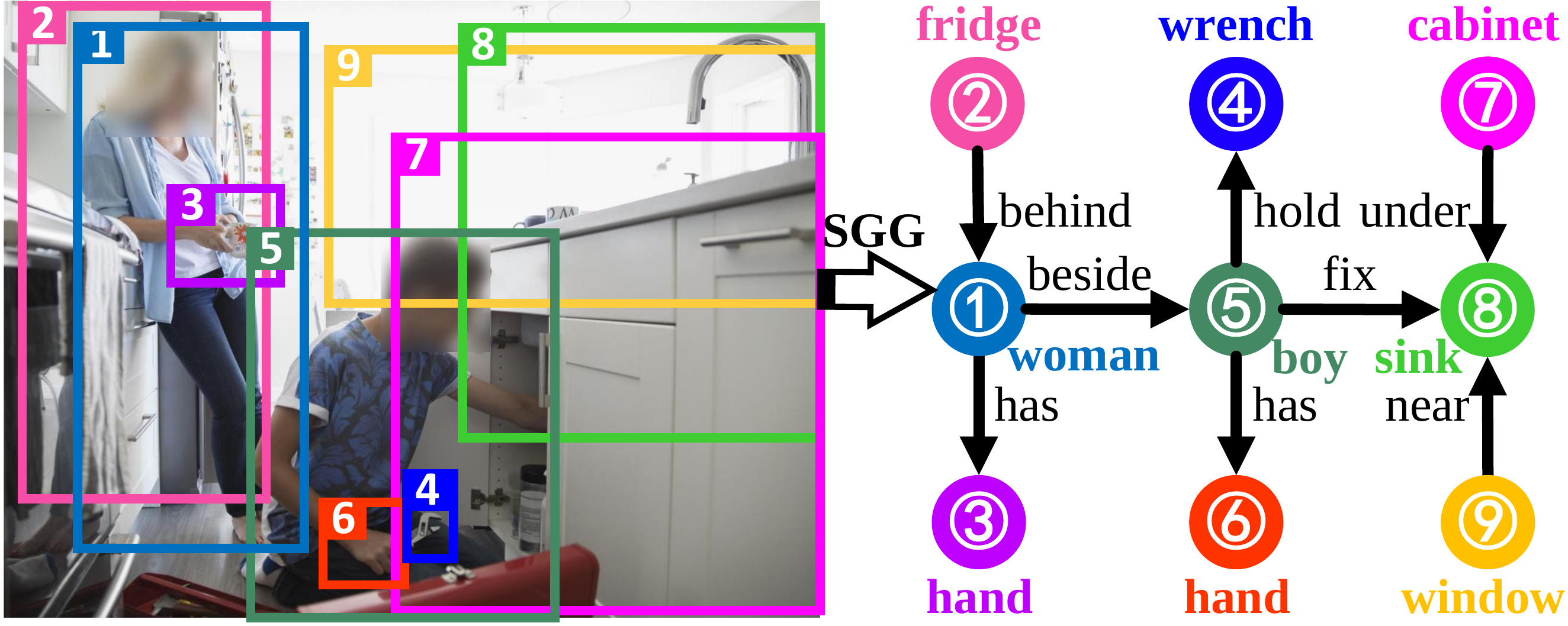}
     \vspace{-5mm}
    \caption{An image and its ground-truth scene graph. Objects and their pairwise relationships are represented as nodes and edges, respectively. Best viewed in color.}
    \vspace{-4mm}
    \label{intro}
\end{figure}

Scene graph generation (SGG) has recently attracted increasing attention from the research community. As illustrated in Figure~\ref{intro}, a visual scene could be depicted in the form of a graph structure, where objects and their pairwise relationships are represented by the nodes and edges, respectively. A triplet constructed by two objects and their corresponding relationship then takes the form  \textit{Subject-Predicate-Object}. Intuitively, the key to SGG methods is to model and explore the connections between objects, as well as those between relationships. Due to their remarkable ability to model the connections between graph components, graph neural networks (GNNs) have been widely adopted in SGG tasks~\cite{yang2018graph,yang2021probabilistic,li2021bipartite,chen2019knowledge,lin2020gps}. 

Despite their effectiveness, existing GNN-based SGG methods only assume homophily \cite{ma2021homophily} between objects/relationships; in other words, these methods calculate the correlations between objects/relationships by implicitly treating all objects/relationships as belonging to the same categories. However, as Figure~\ref{intro} demonstrates, scene graphs fall naturally into the category of heterophilic graphs. We, therefore, argue that heterophily, \ie, the interaction between objects/relationships from different categories, should be modeled directly. In this paper, we focus on the heterophily in class labels, following the definition provided in~\cite{zhu2020beyond, yan2021two}.

Exploring heterophily in the SGG is non-trivial. There are at least two problems that must be considered. First, heterophily in both objects and relationships should be taken into account; however, no prior SGG works have explicitly considered heterophily, and no heterophilic GNNs have exploited heterophily in visual relationships. Second, two objects/relationships characterized by significant occlusion usually have similar visual appearances, despite being from different classes, which increases the difficulty of distinguishing heterophily from homophily. In the light of the above scene graph analysis, in this paper, we propose a Heterophily Learning Network (HL-Net) for SGG to comprehensively and efficiently explore the heterophily in objects/relationships. To the best of our knowledge, HL-Net is the first work to consider heterophily for the SGG. The main contributions of HL-Net are summarized below.

We first propose an \textbf{Adaptive Reweighting Transformer} (ART), which refines object representation with heterophily considered. In more detail, we arrange the pre-layer normalization \cite{xiong2020layer}, residual connection, and feedforward network to deepen the layers and enhance the object feature with contextual information. Furthermore, the refined object representations of different ART layers are aggregated with learnable weights. These weights depend on the contributions of different ART layers and can be both positive and negative. This aggregation procedure is similar to general polynomial graph filtering \cite{shuman2013emerging}, which is naturally able to deal with both the high-frequency context (\ie, heterophily) and low-frequency context (\ie, homophily) between objects \cite{chien2021adaptive}.

We then develop a \textbf{Relationship Feature Propagation} (RFP) module that explores the connections between heterophilic relationships. Two challenges emerge in the design of RFP: the effectiveness of feature propagation and heterophily modeling. To reduce computational complexity, we only require each relationship to contact neighboring relationships that share the subject or object. Moreover, the contextual coefficients obtained from ART are adopted to represent the correlations between relationships. To propagate heterophilic features between relationships, we extend the PageRank-based GNN \cite{klicpera2018predict} to a high-pass graph filter. This approach enables the RFP module to learn the relationship correlation of disparate classes by passing relevant high-frequency graph signals (\ie, heterophily).

Finally, we devise a \textbf{Heterophily-aware Message Passing} (HMP) scheme to identify the heterophily and homophily between objects or relationships in complicated visual scenes (\eg, overlapping objects that belong to different classes). More specifically, HMP utilizes the spatial and visual information of object/relationships to produce signed messages, which can subsequently be applied to adjust the contextual coefficients and guide the learning processes in both ART and RFP.

We conduct extensive experiments on two public datasets: Visual Genome (VG) \cite{krishna2017visual} and Open Images (OI) \cite{kuznetsova2020open}. Experimental results demonstrate that the proposed HL-Net consistently achieves top-level performance. Ablation studies further verify both the necessity and effectiveness of considering heterophily for SGG.

\section{Related Works}
\textbf{Scene Graph Generation.} Early SGG works \cite{zhang2017relationship,zhang2017visual,zhuang2017towards,zhan2019exploring,zhan2020multi} tended to detect each object/relationship independently, ignoring the intrinsic connections between objects/relationships. Recent SGG methods \cite{yang2018graph,tang2019learning,lin2020gps,chen2019knowledge,yang2021probabilistic,li2021bipartite,xu2017scene,zellers2018neural} typically explore visual-contextual information between objects. These methods can be roughly divided into two categories: Recurrent Neural Network (RNN)-based methods and Graph Neural Network (GNN)-based methods. The first category utilizes RNN to encode contextual information. For example, Zeller \etal \cite{zellers2018neural} and Tang \etal \cite{tang2019learning} employed a bidirectional long short-term memory (LSTM) module and a tree structure-based LSTM module to refine object representation using context information, respectively. However, RNN-based SGG approaches may not adequately depict connections between distant objects. The second category of methods utilizes GNN to propagate contextual information. For example, Yang \etal \cite{yang2018graph} proposed an attentional graph convolutional network to refine object and relationship representations, while Lin \etal \cite{lin2020gps} proposed a direction-aware message-passing module that encodes the edge direction information. However, recent studies \cite{zhu2020beyond,yan2021two} have proven that most existing GNN-based methods struggle to describe connections under heterophily. Unfortunately, scene graphs are naturally heterophilic. To address these problems, we propose HL-Net in an attempt to capture the heterophilic property of scene graphs. To the best of our knowledge, HL-Net is the first work to explicitly consider heterophily in the SGG. 

\begin{figure*}[tbp]
  \begin{center}
    \includegraphics[width=1\linewidth, height=.43\linewidth]{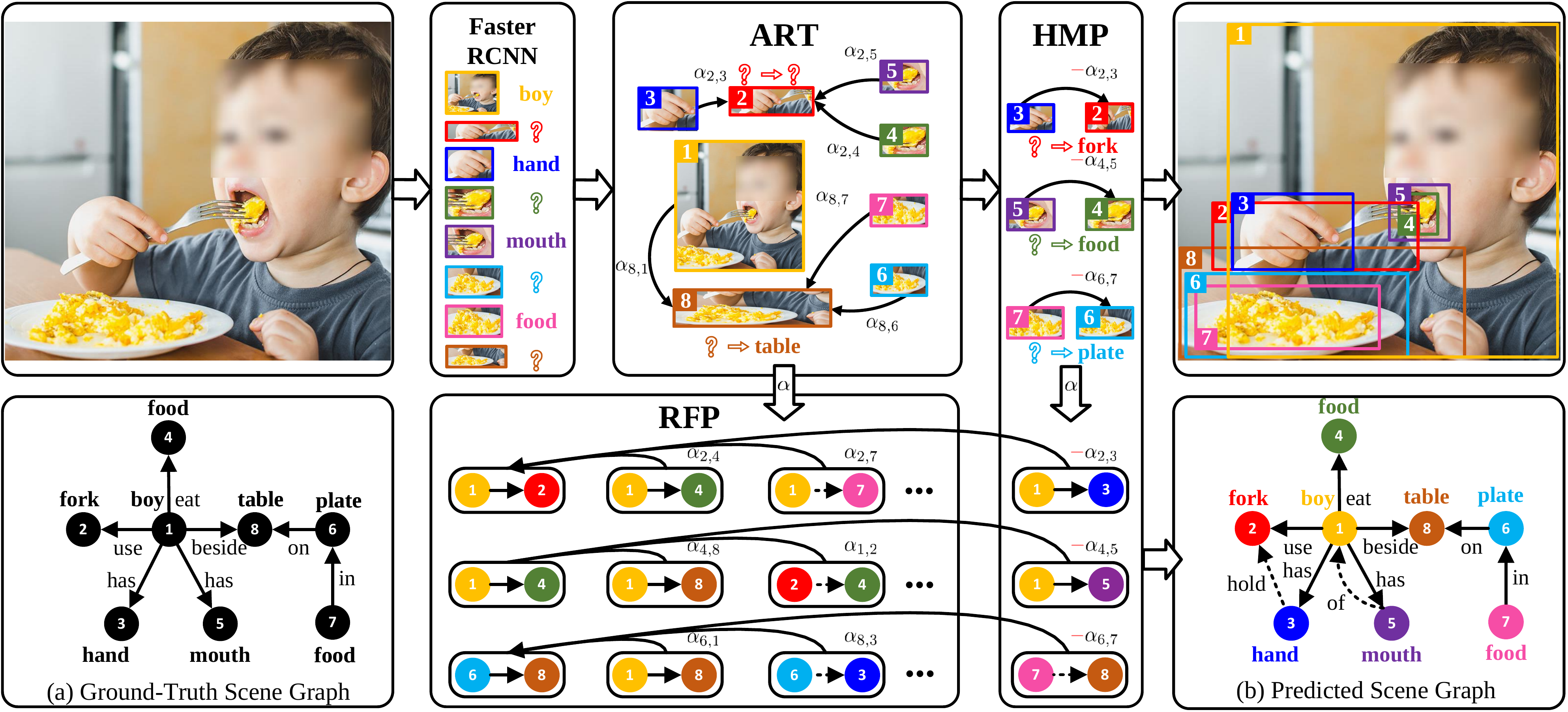}
  \end{center}
  \vspace{-5mm}
  \caption{The framework of HL-Net. HL-Net obtains object proposals through Faster R-CNN \cite{ren2015faster}. It then improves the performance of SGG through the application of two novel modules: (1) an ART module that enables message-passing between objects with heterophily considered; (2) an RFP module that explores connections between heterophilic relationships. Moreover, HL-Net includes an HMP scheme that identifies the heterophily and homophily between objects and those between relationships under complicated visual scenes.}
  \label{framework}
\end{figure*}

\textbf{GNNs and Heterophily.}

Recent works \cite{liu2020non,pei2019geom,ma2021homophily,zhu2020beyond} have shown that the use of certain GNNs (\eg, Graph Convolutional Network \cite{kipf2016semi} and Graph Attention Network \cite{velivckovic2017graph}) can lead to significant performance loss in heterophilous settings. A number of works have attempted to address this issue. For example, Zhu \etal \cite{zhu2020beyond} proposed a set of designs including embedding separation, higher-order neighborhoods aggregation, and intermediate representations that enable GNN to perform well under heterophilic settings. Zhou \etal \cite{zhou2020understanding} introduced a new belief propagation-based GNN model. Chien \etal \cite {chien2021adaptive} devised a generalized PageRank-based GNN architecture that adaptively learns the propagation weights to determine the polynomial graph filter for heterophilic graph. Yan et al \cite{yan2021two} proposed a model that allows negative interactions between nodes in order to capture heterophily. However, the above approaches focus primarily on the task of natural language processing, (\eg, node classification for citation graphs). Therefore, applying the above-mentioned heterophilic GNNs directly to the SGG may not adequately solve the heterophily problem for visual content.

\section{Heterophily Learning Network}
This section presents the framework of our Heterophily Learning Network (HL-Net). As Figure~\ref{framework} illustrates, HL-Net comprises an adaptive reweighting transformer (ART) module and a relationship feature propagation (RFP) module. The ART module strengthens the network's object classification ability by means of  heterophily-aware message passing between object representations. The RFP module promotes its predicate classification performance by exploring connections between heterophilic relationships. We further propose a heterophily-aware message passing (HMP) scheme that identifies the heterophily and homophily between objects, along with those between relationships, and enhance the power of both ART and RFP. In the below, we will describe these three components sequentially.

\subsection{Preliminary}
\noindent \textbf{Notations}. We first introduce the notations used in this section. We adopt exactly the same approach used in \cite{zellers2018neural} to obtain the representation ${\emph {\textbf{x}}_i}$ for the $i$-th object/node. More specifically, ${\emph {\textbf{x}}_i}$ is transformed using linear projection from the concatenation of the visual appearance feature, object classification probabilities, and the spatial feature. We extract appearance feature form the union box of the two nodes $i$ and $j$, denoted as ${\emph {\textbf{x}}_{ij}}$; similarly, the appearance feature for the union box of three nodes $i$, $j$, and $k$ is denoted as ${\emph {\textbf{x}}_{ijk}}$. The relationship feature between the $i$-th and $j$-th nodes is represented as ${\emph {\textbf{r}}_{ij}}$, where the $i$-th node is the {\textit{subject}} and the $j$-th node is the {\textit{object}}. ${\emph {\textbf{B}}_{ij}}$ is used to represent the relative spatial feature between the $i$-th and $j$-th nodes. It is obtained by applying two convolutional layers and two FC layers to binary maps of size $14\times 14\times 2$, with each channel representing the area of one node. Similarly, ${\emph {\textbf{B}}_{ij,k}}$ denotes the relative spatial feature between the union box of nodes $i$, $j$ and the bounding box for the node $k$, and ${\emph {\textbf{B}}_{i,jk}}$ represent the relative spatial feature between the bounding box of node $i$ and the union box of nodes $j$, $k$. $\odot$ represents the Hadmard product. ${\cal N}_i$ denotes the set of neighboring objects for the $i$-th node. ${\cal N}_{{{\emph {\textbf{r}}}_{ij}}}$ indicates the set of neighboring relationships of ${{\emph {\textbf{r}}}_{ij}}$. Finally, ${\emph {\textbf{W}}}$ stands for a linear transformation matrix and ${\emph {\textbf{w}}}$ means a linear projection vector. 

\noindent \textbf{Homophily and Heterophily}. 
Given a set of node classes, {\textit{homophily}} describes the tendency of a node to have the same class as its neighbors, while {\textit{heterophily}} describes the tendency of a node to have different classes as its neighbors. In more detail, \cite{pei2019geom,yan2021two,chien2021adaptive} proposed a metric for measuring the level of homophily of nodes in a graph: $h=\frac{1}{\left\| {{\cal V}} \right\|} \sum_{i \in {\cal V}} \frac{{\left\| {{\cal N}^s_i} \right\|}}{{\left\| {{\cal N}_i} \right\|}}$, here $\cal V$ represents a node set, ${{\cal N}^s_i}$ denotes the set of neighboring nodes with the same label as the $i$-th node, and $\left\|\cdot \right\|$ is the cardinality operator. Accordingly, ${h}\rightarrow$1 corresponds to strong homophily, while ${h}\rightarrow$0 indicates strong heterophily. This definition could be extended to describe the homophily and heterophily of edges.

\subsection{Adaptive Reweighting Transformer}

Graph attention network has been widely adopted in existing SGG methods \cite{yang2018graph,chen2019knowledge,lin2020gps,li2021bipartite}. However, recent works \cite{zhu2020beyond,chien2021adaptive,yan2021two} have shown that graph attention network implicitly assumes homophily between nodes; therefore and accordingly ignores the property of heterophily in scene graph. To address this problem, we propose the ART module, which includes two components: namely, the {\bf Pre-LN Transformer} and {\bf Adaptive Graph Filter}.


\noindent\textbf{Pre-LN Transformer}:
We adopt the same approach as in \cite{lin2020gps} to obtain the contextual coefficient $c_{ij}$ between two nodes $i$ and $j$ as follows:
\begin{equation}\label{spa}
c_{ij}={\emph {\textbf{w}}}^{T}_c({\emph {\textbf{W}}}_{c1} {\emph {\textbf{x}}}_{i}  \odot  {\emph {\textbf{W}}}_{c2} {\emph {\textbf{x}}}_{j}  \odot ({\emph {\textbf{x}}}_{i j}+{\emph {\textbf{B}}}_{ij})).
\end{equation}

Inspired by \cite{xiong2020layer}, we employ pre-layer normalization (Pre-LN) to stabilize the model training. The neighboring messages for the $i$-th node can be aggregated as follows:
\begin{equation}\label{F}
{\cal F}({\cal N}_i) = \sum\nolimits_{j \in {{\cal N}_i}} {\alpha _{ij}\sigma ( {{\emph {\textbf{W}}_{\cal F}}{\mathop{\rm LN}}( {{\emph {\textbf{x}}}_j} )} )},
\end{equation}
where $\sigma$ denotes the ReLU activation function, while $\alpha_{i j}$ is a contextual coefficient, which is obtained by normalizing $c_{ij}$ with softmax. Furthermore, we adopt layer normalization \cite{ba2016layer}, FFN layer \cite{vaswani2017attention}, and residual connection sequentially to refine the node representations. Consequently, the output of the $u$-th layer for the $i$-th node can be denoted as follows:
\begin{equation}\label{node_up}
{{\emph {\textbf{x}}}_i^{u + 1}} = {{{\emph {\textbf{z}}}_{i}^u}} + \mathop{\rm FFN}( {\mathop{\rm LN}}(\overbrace{{{\emph {\textbf{x}}}_i^{u}}+{\cal F}^u({\cal N}_i)}^{{{\emph {\textbf{z}}}_{i}^u}})).  
\end{equation}

\noindent\textbf{Adaptive Graph Filter}:
As proven in \cite{zhu2020graph}, existing GNN approaches \cite{klicpera2018predict,xu2018representation,wu2019simplifying} typically focus on emphasizing homophily by aggregating the outputs of different GNN layers with non-negative weights. This aggregation step can be understood as a low-pass graph filter that emphasizes the low-frequency part of the graph signal (\ie, homophily). However, this filter suppresses high-frequency components (\ie, heterophily) in the graph signal. In comparison, if the outputs of GNN layers can be aggregated with negative weights, a polynomial graph filter \cite {shuman2013emerging} for heterophilic graphs can be obtained \cite{chien2021adaptive}. 

Motivated by the above analysis, ART calculates the final node representation as follows:
\begin{equation}\label{out}
{\hat {\emph {\textbf{x}}}_i} ={\mathop{\rm LN}}(\sum\nolimits_{u=1}^{U} \gamma_{u} {\emph {\textbf{x}}}_{i}^{u}),
\end{equation}
Here, $U$ denotes the number of GNN layers while $\gamma_u$ represents the weight of the $u$-th GNN layer. Note that $\gamma_u$ can be a negative number and is optimized simultaneously with the whole HL-Net in an end-to-end manner. To properly capture the heterophilic property of the scene graph, we heuristically initialize $\gamma_u$  with a high-pass filter based formulation (the proof of which is provided in Appendix C.1.2), as follows:
\begin{equation}\label{gamma}
{\gamma_u} = \frac{{{{( - \tau )}^{u - 1}}}}{{\sum\nolimits_{u = 1}^U {| {{{( - \tau )}^{u - 1}}} |} }}, 
\end{equation}
where $\tau \in(0,1)$ is a hyperparameter. More details regarding the initialization of ${\gamma_u}$ can be found in the appendix.

Finally, the classification score vector of the $i$-th node can be obtained as follows: ${\emph {\textbf{v}}}_{i}={\rm softmax}({\emph {\textbf{W}}}_{v}\hat{\emph {\textbf{x}}}_{i})$. Comparisons between ART and existing message passing modules can be found in the supplementary materials.

\subsection{Relationship Feature Propagation}\label{rlp}

Existing SGG works typically ignore correlations between relationships. In this subsection, we propose the RFP module to use the inter-relationship connections under heterophilic settings. To the best of our knowledge, no existing heterophilic GNNs have explicitly explored the connections between edges.

An intuitive design choice for RFP is to use the same architecture as ART. However, there are $N(N-1)$ potential relationships for an image containing $N$ objects, implying that using the same structure as ART for RFP incurs a high computational cost. Moreover, there are no meaningful relationships between the majority of object pairs. To address the above issues, we adopt two strategies. First, we only model the connections between relationships that share the same \textit{subject} or \textit{object}. Second, we utilize the message-passing coefficients between nodes to guide edges since the connections between relationships can be decoupled into connections between their related objects.

Specifically, the representation of one relationship ${\emph {\textbf{r}}}_{ij}$ is obtained as follows:
 
\begin{equation}\label{r1}
{\emph {\textbf{r}}}_{ij}={\hat {\emph {\textbf{x}}}_i} * {\hat {\emph {\textbf{x}}}_j} * ({{\emph {\textbf{x}}}_{ij}}+{{\emph {\textbf{B}}}_{ij}}),
\end{equation}
where $*$ denotes a fusion function defined in \cite{tang2019learning}: ${\emph {\textbf{x}}} * {\emph {\textbf{y}}} = {\mathop{\rm ReLU}\nolimits} \left( {{{\emph {\textbf{W}}}_x}{\emph {\textbf{x}}} + {{\emph {\textbf{W}}}_y}{\emph {\textbf{y}}}} \right) - \left( {{{\emph {\textbf{W}}}_x}{\emph {\textbf{x}}} - {{\emph {\textbf{W}}}_y}{\emph {\textbf{y}}}} \right) \odot \left( {{{\emph {\textbf{W}}}_x}{\emph {\textbf{x}}} - {{\emph {\textbf{W}}}_y}{\emph {\textbf{y}}}} \right)$.
We then obtain the initial classification score vector of the relationship between the $i$-th and $j$-th nodes as follows:
\begin{equation}
{{\emph {\textbf{p}}}^{0}_{ij}}={{\emph {\textbf{W}}}_{p}}\sigma({{\emph {\textbf{W}}}_{r}}{{\emph {\textbf{r}}}_{ij}}).
\end{equation}

Subsequently, we obtain the messages passed from neighboring relationships to ${{\emph {\textbf{r}}}_{ij}}$ as follows:
\begin{equation}\label{msg_rel111}
{\cal H}({\cal N}_{{\emph {\textbf{r}}}_{ij}} )=\sum\nolimits_{l \in {{\cal N}_j}} {\hat \alpha_{jl} { {{\emph {\textbf{p}}}_{il}} } }+  \sum\nolimits_{m \in {{\cal N}_i}} {\hat \alpha_{im} { {{\emph {\textbf{p}}}_{mj}} } },
\end{equation}
where $\hat \alpha_{jl}$ and $\hat \alpha_{im}$ indicate the normalized contextual coefficient according to the elements in ${\cal N}_j+{\cal N}_i$.

To reduce the computational complexity, some GNN models \cite {klicpera2018predict, li2019optimizing} have utilized PageRank-based approaches to propagate the label information. However, as proven in \cite{chien2021adaptive}, these methods act as low-pass graph filters, which invariably suppress the high-frequency component, namely the heterophily. To address this issue, we formulate the output of the $n$-th propagation layer for the relationship ${{\emph {\textbf{r}}}_{ij}}$ as follows:
\begin{equation}\label{up}
{{\emph {\textbf{p}}}^{k+1}_{ij}} = \beta ({ {{\emph {\textbf{p}}}^{k}_{ij}} }+{\cal H}^{k}({\cal N}_{{{\emph {\textbf{r}}}_{ij}}} ) )+ (1-\beta) {{\emph {\textbf{p}}}^{0}_{ij}}.
\end{equation}
Here, $\beta$ indicates the teleport probability \cite{chien2021adaptive}, which controls how fast Eq.~\eqref{up} moves away from ${{\emph {\textbf{p}}}^{0}_{ij}}$. As described in Theorem 4.1 of \cite{chien2021adaptive}, Eq.~\eqref{up} could be considered to operate as a high-pass graph filter such that it allows the teleport probability $\beta$ to be negative. In other words, Eq.~\eqref{up} enables the model to pass relevant high-frequency graph signals, such as heterophily.

Finally, the classification score vector for the relationship between the $i$-th and $j$-th nodes can be written as follows:
\begin{equation}
{{\emph {\textbf{t}}}_{ij}}= {\rm softmax}({{\emph {\textbf{p}}}^{K}_{ij}}+{{\emph {\textbf{f}}}_{ij}}),
\end{equation}
Here, $K$ denotes the number of RFP layer. ${{\emph {\textbf{f}}}_{ij}}$ indicates the relationship distribution vector between the object categories of the $i$-th and $j$-th nodes in the training set. It functions in the same way as frequency bias and has been widely adopted in existing works~\cite{zellers2018neural,lin2020gps,yang2021probabilistic,li2021bipartite}.

\subsection{Heterophily-aware Message Passing}

\begin{table*}[t]
\small
\setlength{\tabcolsep}{1.1mm}
\centering
\begin{tabular}{l|l|clclc|c lclc|clclc|l}
\hline
                        &                                       & \multicolumn{5}{c|}{SGDET}                                                                                             & \multicolumn{5}{c|}{SGCLS}                                                                                                         & \multicolumn{5}{c|}{PREDCLS}                                                                                                                      \\
           Backbone                  & Model                               & R@20                   & \multicolumn{1}{c}{} & R@50                   & \multicolumn{1}{c}{} & R@100                  & R@20                   & \multicolumn{1}{c}{}       & R@50                   & \multicolumn{1}{c}{}       & R@100                  & R@20                   & \multicolumn{1}{c}{}       & R@50                   & \multicolumn{1}{c}{}       & R@100    & Mean                                \\ \hline \hline

                          & IMP$^\diamond$  \cite{xu2017scene}                     & 14.6                   &                      & 20.7                   &                      & 24.5                   & 31.7                   &                            & 34.6                   &                            & 35.4                   & 52.7                   &                            & 59.3                   &                            & 61.3    & \multicolumn{1}{c}{39.3}             \\
                          & MOTIFS$^\diamond$  \cite{zellers2018neural}                    & 21.4                   &                      & 27.2                   &                      & 30.3                   & 32.9                   &                            & 35.8                   &                            & 36.5                   & 58.5                   &                            & 65.2                   &                            & 67.1                               &   \multicolumn{1}{c}{43.7}     \\
      & KERN$^\diamond$  \cite{chen2019knowledge}                     & -                      &                      & 27.1                   &                      & 29.8                   & -                      &                            & 36.7                   &                            & 37.4                   & -                      &                            & 65.8                   &                            & 67.6              & \multicolumn{1}{c}{44.1}       \\
 & GPI$^\diamond$  \cite{herzig2018mapping}                     & -                      &                      & -                  &                      & -                   & -                      &                            & 36.5                  &                            & 38.8                  &  -                      &                            & 65.1                 &                           & 66.9           &     \multicolumn{1}{c}{-}       \\

                    & VCTREE$^\diamond$ \cite{tang2019learning}                 & 22.0                   &                      & 27.9                   &                      & 31.3                   & 35.2                   &                            & 38.1                   &                            & 38.8                   & 60.1                   &                            & 66.4                   &                            & 68.1                        &     \multicolumn{1}{c}{45.1}                \\
    \multicolumn{1}{l|}{VGG-16}                          &  GPS-Net$^\diamond$ \cite{lin2020gps} & 22.6 &                      & 28.4 &                      &  31.7 &  36.1 &                            &  39.2 &                            &  40.1 &  60.7 &                            &  66.9 &                            &  68.8 &     \multicolumn{1}{c}{45.9} \\

                                                    &  R-CAGCN$^\diamond$ \cite{yang2021probabilistic} & 22.1 &                      & 28.1 &                      &  31.3 &  35.4 &                            &  38.3 &                            &  39.0&  60.2 &                            &  66.6 &                            &  68.3 &     \multicolumn{1}{c}{45.3}\\
                          &\bf HL-Net$^\diamond$                  &\bf 22.9                      & \multicolumn{1}{c}{} &              28.5           & \multicolumn{1}{c}{} &                  31.9      & 37.2                  & \multicolumn{1}{c}{}       & 39.8                   & \multicolumn{1}{c}{}       & 40.8                   & 61.3                  & \multicolumn{1}{c}{}       & 67.5                   & \multicolumn{1}{c}{}       & 69.5                 &     \multicolumn{1}{c}{ 46.3}  \\\cline{2-18} 
                             &  RelDN$^\ddagger$  \cite{zhang2019graphical} & - &                      &-  &                      &  32.7  &  - &                            &  - &                            &  36.8 &  - &                            &  - &                            &  68.4 &     \multicolumn{1}{c}{-}\\
 
                            &  Seq2Seq-RL$^\ddagger$  \cite{lu2021context} & 22.1 &                      & 30.9 &                      &  34.4  &  34.5 &                            &  38.3 &                            &  39.0 &  60.3 &                            &  66.4 &                            &  68.5 &     \multicolumn{1}{c}{46.3}\\
                                                    &\bf HL-Net $^\ddagger$                 & 22.5                     & \multicolumn{1}{c}{} &             \bf 31.3           & \multicolumn{1}{c}{} &                 \bf 34.7      &\bf 37.4                  & \multicolumn{1}{c}{}       &\bf 40.4                   & \multicolumn{1}{c}{}       &\bf 41.3                   &\bf 61.6                  & \multicolumn{1}{c}{}       &\bf 67.7                   & \multicolumn{1}{c}{}       &\bf 69.7                 &     \multicolumn{1}{c}{\bf 47.5}  \\\hline
                                               
 & VTransE \cite{tang2020unbiased}                     & 23.0                   &                      & 29.7                   &                      & 34.3                   & 35.4                   &                            & 38.6                   &                            & 39.4                   & 59.0                   &                            & 65.7                   &                            & 67.6         &  \multicolumn{1}{c}{45.9}            \\

& VCTREE \cite{tang2019learning}                         & 24.7                   &                      & 31.5                   &                      & 36.2                   & 37.0                   &                            & 40.5                   &                            & 41.4                   & 59.8                   &                            & 66.2                  &                            & 68.1                                      &   \multicolumn{1}{c}{47.3} \\

 \multicolumn{1}{c|}{RX-101} & MOTIFS \cite{zellers2018neural}                      & 25.1                   &                      & 32.1                  &                      & 36.9                   & 35.8                   &                            & 39.1                   &                            & 39.9                   & 59.5                   &                            & 66.0                   &                            & 67.9                              & \multicolumn{1}{c}{47.0}       \\
 
  & SGGNLS \cite{zhong2021learning}                         & 24.6                   &                      & 31.8                  &                      & 36.3                   & 36.5                   &                            & 40.0                   &                            & 40.8                   & 58.7                   &                            & 65.6                   &                            & 67.4                              & \multicolumn{1}{c}{47.0}       \\

                          & \bf HL-Net             &\bf 26.0  &                      & \bf 33.7   &                      &\bf 38.1    & \bf 38.8                   & \multicolumn{1}{c}{}       &\bf 42.6                  & \multicolumn{1}{c}{}       &\bf 43.5                    &\bf 60.7                  & \multicolumn{1}{c}{}       &\bf 67.0                   & \multicolumn{1}{c}{}       &\bf 68.9           & \multicolumn{1}{c}{\bf 49.0}           \\ 
                          \hline
\end{tabular}
\caption{Performance comparisons with state-of-the-art methods on the VG dataset. We compute the mean on all tasks over R@50 and R@100. $^\diamond$ and $^\ddagger$ denote the methods using the same Faster-RCNN detector as \cite{zellers2018neural} and \cite{zhang2019graphical}, respectively.}
\label{vg1}
\end{table*}

Heterophily causes GNNs to experience performance degradation. Recent works \cite{yan2021two} in GNN architecture design mitigate this problem by allowing the messages from inter-class neighbors to be multiplied by a negative sign. This operation enables the mean distance between inter-class nodes to be less affected in the aggregation procedure. To better distinguish between the homophily and heterophily in the scene graph, especially within the complicated visual scene (\ie, occlusion), we define a sign function to adjust the non-negative contextual coefficient between nodes or edges. Furthermore, this sign function indicates whether they belong to the same category. In more detail, we formulate the sign message between two nodes with the features of their union box, defined as follows:
\begin{equation}\label{s_ij}
{s_{ij}} = {\rm tanh}({{\emph {\textbf{w}}}^{T}_{s}}\sigma({{\emph {\textbf{W}}}_{s}}[{{\emph {\textbf{x}}}_{ij}}+{{\emph {\textbf{B}}}_{ij}},{{\emph {\textbf{v}}}_{i}},{{\emph {\textbf{v}}}_{j}}])),
\end{equation}
where $[,]$ represents the concatenation operation. Tanh is utilized to approximate the sign function and has the additional benefit of being differential. In the training process, a binary cross-entropy (BCE) loss is utilized for supervision with ground-truth sign labels $y^s_{ij}\in \{ -1,1\}$; here, 1 and -1 indicate that the two nodes belong to the same and different object categories, respectively. By integrating the sign information into Eq.~\eqref{F}, the message between two nodes can be refined as follows:

\begin{equation}\label{F1}
{\cal F}({\cal N}_i) = \sum\nolimits_{j \in {{\cal N}_i}} {s_{ij}\alpha_{ij}\sigma ( {{\emph {\textbf{W}}_{\cal F}}{\mathop{\rm LN}}( {{\emph {\textbf{x}}}_j} )} )}. 
\end{equation}



Similarly, the sign function for two neighboring edges with the same {\textit{subject}} or {\textit{object}} can be approximated as follows:
\begin{equation}\label{r1}
\begin{aligned}
q_{{il}\rightarrow{ij}}&={\rm tanh}({{\emph {\textbf{w}}}^{T}_{q}}({\hat {\emph {\textbf{x}}}_i} * {{\emph {\textbf{x}}}_{lj}} * ({{\emph {\textbf{x}}}_{ijl}}+{{\emph {\textbf{B}}}_{i,lj}})))\\
q_{{mj}\rightarrow{ij}}&={\rm tanh}({{\emph {\textbf{w}}}^{T}_{q}}({ {\emph {\textbf{x}}}_{im}} *   {\hat {\emph {\textbf{x}}}_j} *  ({{\emph {\textbf{x}}}_{ijm}}+{{\emph {\textbf{B}}}_{im,j}}))),
\end{aligned}
\end{equation}
where $q_{{il}\rightarrow{ij}}$ denotes two edges ${\emph {\textbf{r}}_{ij}}$ and ${\emph {\textbf{r}}_{il}}$ that share the same {\textit{subject}}, while $q_{{mj}\rightarrow{ij}}$ indicates two edges ${\emph {\textbf{r}}_{ij}}$ and ${\emph {\textbf{r}}_{mj}}$ that share the same {\textit{object}}.

A BCE loss is adopted for the supervision on $q_{{il}\rightarrow{ij}}$ and $q_{{mj}\rightarrow{ij}}$, respectively. The ground-truth labels are 1 and -1 for two edges that belong to the same and different categories, respectively.
Finally, the sign information is utilized to refine the messages between neighboring edges defined in Eq.~\eqref{msg_rel111} as follows:

\begin{equation}\label{msg_rel}
{\cal H}({\cal N}_{{\emph {\textbf{r}}}_{ij}} )= \sum_{l \in {{\cal N}_j}} {q_{{il}\rightarrow{ij}}\hat \alpha_{jl} { {{\emph {\textbf{p}}}_{il}} } }+ \sum_{m \in {{\cal N}_i}} {q_{mj\rightarrow{ij}}\hat \alpha_{im} { {{\emph {\textbf{p}}}_{mj}} } }.
\end{equation}

\subsection{SGG by HL-Net}
During training, the overall loss function $\mathcal L$ of HL-Net can be expressed as follows:
\begin{equation}
{\mathcal L}= {\mathcal L}_v + {\mathcal L}_e + {\mathcal L}^{v}_{bce} + {\mathcal L}^{e}_{bce},
\end{equation}
where ${\mathcal L}_v$ and ${\mathcal L}_e$ are the standard cross-entropy loss for object and relationship classification, respectively. Moreover, ${\mathcal L}^{v}_{bce}$ and ${\mathcal L}^{e}_{bce}$ represent the BCE loss for sign prediction in object and relationship classification, respectively.

During testing, the object category for the $i$-th node is predicted by the following equation:
\begin{equation}
o_i = {\arg{\max} _{o \in {\mathcal O}}}({{\emph {\textbf{v}}}_{i}(o)}),
\end{equation}
where $\mathcal O$ represents the set of object categories. The relationship category of the edge between the $i$-th and $j$-th nodes can be obtained as follows:
\begin{equation}
e_{ij}={\arg{\max} _{r \in {\mathcal R}}}({{\emph {\textbf{t}}}_{ij}(r)}),    
\end{equation}
where $\mathcal R$ represents the set of relationship categories.

\section{Experiments}\label{SecExp}

\subsection{Dataset and Evaluation Settings}
\noindent\textbf{Visual Genome}: We follow the same data cleaning strategy \cite{xu2017scene} that has been widely adopted in several recent works. More specifically, the most frequently occurring 150 object categories and 50 relationship categories are utilized for evaluation. The scene graph for each image consists of 11.6 objects and 6.2 relationships on average. We follow three conventional protocols for evaluation: 1) Scene Graph Detection (SGDET): Given an image, the model detects object bounding boxes and predicts both the object and relationship categories for each bounding box pair. 2) Scene Graph Classification (SGCLS): Given the ground-truth location of object bounding boxes, the model predicts both the object and relationship categories. 3) Predicate Classification (PREDCLS): Given the ground-truth object bounding boxes and their object categories, the model predicts only the relationship categories. All three settings are evaluated according to Recall@$K$ (R@$K$) metrics, where $K$ is set to 20, 50, and 100, respectively. 

\noindent\textbf{Open Images}: Open Images (OI) \cite{kuznetsova2020open} is a large-scale dataset proposed by Google. We conduct our experiments on Open Images V4 and V6. In more detail, the Open Images V4 dataset contains 53,953 and 3,234 images as the training and validation sets, respectively. It comprises a total of 57 object categories and 9 predicate categories. Open Images V6 contains 126,368/1,813/5,322 images used for training/validation/testing, respectively. It has 301 object categories and 31 predicate categories. We follow the same data processing and evaluation protocols outlined in \cite{zhang2019graphical,lin2020gps,li2021bipartite}. More specifically, the results are evaluated by calculating Recall@50 (R@50), the weighted mean AP of relationships $({\rm {wmAP}}_{rel})$, and the weighted mean AP of phrase (${\rm {wmAP}}_{phr}$). The final score is given by score$_{wtd}=0.2\times {\rm R}@50+0.4\times {\rm {wmAP}}_{rel}+0.4\times {\rm {wmAP}}_{phr}$. Note that ${\rm {wmAP}}_{rel}$ evaluates the AP of the predicted triplet in which both the subject and object boxes have an IoU of at least 0.5 with ground truth, while ${\rm {wmAP}}_{phr}$ evaluates the AP of the predicted triplet where the union area of the subject and object boxes has an IoU of at least 0.5 with ground truth.

\subsubsection{Implementation Details}
To facilitate a fair comparison with the majority of existing works, we utilize ResNeXt-101-FPN \cite{lin2017feature, xie2017aggregated} as the backbone for the OI database. We adopt both ResNeXt-101-FPN \cite{lin2017feature, xie2017aggregated} and VGG-16 \cite{simonyan2014very} as the backbones for the VG database. During training, we freeze the layers before the ROIAlign layer and optimize the remaining layers in the model using both the object and relationship classification losses. We optimize HL-Net via Stochastic Gradient Descent (SGD) with momentum, using an initial learning rate of 10$^{-3}$ and a batch size of 6. The top-64 object proposals in each image are selected following per-class non-maximal suppression (NMS) with an IoU of 0.3. Moreover, the sampling ratio between pairs without any relationship (background pairs) and those with relationships during training is set to 3:1. We further set the teleport probability $\beta$ to -0.5.

\subsection{Comparisons with State-of-the-Art Methods}

\noindent\textbf{Visual Genome:} Table~\ref{vg1} shows that HL-Net outperforms all state-of-the-art methods on all metrics. More specifically, HL-Net outperforms one very recent GNN-based SGG model, named R-CAGCN \cite{yang2021probabilistic}, by 1.0$\%$ on average at R@50 and R@100 over the three protocols. It further outperforms R-CAGCN \cite{yang2021probabilistic} by 0.6 $\%$, 1.8 $\%$, and 1.2 $\%$ on SGDET, SGCLS, and PREDCLS at R@100, respectively. Moreover, HL-Net outperforms VCTREE \cite{tang2019learning,tang2020unbiased} using the same ResNeXt-101-FPN backbone by 2.1$\%$ and 1.9$\%$ on SGCLS and SGDET at R@100, respectively.
\begin{table}[]
\small
\setlength{\tabcolsep}{0.6mm}
\centering
\begin{tabular}{l|l|cc|cc}
 \hline
                            &         & \multicolumn{2}{c|}{C-SGCLS} & \multicolumn{2}{c}{S-SGCLS} \\
Backbone                    & Method      & R@50      & R@100         & R@50      & R@100      \\ \hline\hline
                            & MOTIFS \cite{zellers2018neural}       & 32.5      & 33.4        & 35.5      & 36.4       \\
                            & KERN \cite{chen2019knowledge}          & 33.7      & 34.6           & 36.8      & 37.7       \\
                                                        & VCTREE \cite{tang2019learning}       & 34.9      & 35.9            & 38.0      & 38.9       \\ 
VGG-16                     & GPS-Net \cite{lin2020gps}    & 35.8      & 37.1             & 38.4      & 39.3       \\\cline{2-6} 
                            & \bf HL-Net             &\bf 38.3           &\bf  39.4                    & \bf38.7          &\bf 39.6           \\ \hline
                            & VTransE \cite{tang2020unbiased}        &   34.9        & 36.0                 &   38.7        & 39.9        \\
\multicolumn{1}{c|}{RX-101}  & MOTIFS \cite{zellers2018neural}          &   35.4        &  36.5                   &   39.9        &   40.9         \\ \cline{2-6} 
                            & {\bf HL-Net}           & {\bf 41.0}           &  {\bf 42.2}            & {\bf 41.8}          &   {\bf 42.7}         \\
                            \hline
\end{tabular}
{\caption{SGCLS performance comparison under occlusion and non-occlusion scenarios. C-SGCLS denotes the SGCLS performance of triplets where at least one object is heavily occluded by others, otherwise, results are marked S-SGCLS.}\label{crowd}}
\end{table}
\begin{table}[]
\small
\centering
\setlength{\tabcolsep}{1.2mm}
\begin{tabular}{ll|c|c|c}
\hline
& & \multicolumn{1}{c|}{SGDET} & \multicolumn{1}{c|}{SGCLS} & \multicolumn{1}{c}{PREDCLS}  \\
Model & & mR@100 & mR@100 & mR@100 \\ \hline\hline
IMP \cite{xu2017scene} & & 4.8 & 6.0  & 10.5 \\
FREQ \cite{zellers2018neural} && 7.1  & 8.5 & 16.0 \\
MOTIFS \cite{zellers2018neural} && 6.6  & 8.2 & 15.3 \\
KERN \cite{chen2019knowledge}  && 7.3  & 10.0 & 19.2  \\
VCTREE-SL \cite{tang2019learning}  & &7.7   & 10.5  & 18.5 \\
VCTREE-HL \cite{tang2019learning} && 8.0   & 10.8  & 19.4  \\ 
R-CAGCN \cite{yang2021probabilistic} & & 8.8  & 11.1  & 19.9 \\\hline
\bf HL-Net  & &\bf 9.2  &\bf13.5  &\bf 22.8 \\ \hline
\end{tabular}
{\caption{Performance comparison on mean recall $(\%)$ across all 50 relationship categories. All methods in this table adopt the same Faster-RCNN from \cite{zellers2018neural} model as object detector.}\label{tbam}}
\end{table}

\begin{table}[]
\small
\setlength{\tabcolsep}{1.3mm}
\centering
\begin{tabular}{ll|c|cc|c}
\hline
\multicolumn{1}{c|}{\multirow{2}{*}{Dataset}} &
  \multicolumn{1}{c|}{\multirow{2}{*}{Model}} &
  \multicolumn{1}{c|}{\multirow{2}{*}{R@50}} &
  \multicolumn{2}{c|}{WmAP} &
  \multicolumn{1}{c}{\multirow{2}{*}{${\rm {score}}_{wtd}$}} \\ \cline{4-5}
\multicolumn{1}{c|}{} &
  \multicolumn{1}{c|}{} &
  \multicolumn{1}{c|}{} &
  rel &
  \multicolumn{1}{l|}{phr} &
  \multicolumn{1}{c}{} \\ \hline \hline
\multicolumn{1}{l|}{}                  & RelDN \cite{zhang2019graphical}      & 74.9                & 35.5                & 38.5                & 44.6                 \\
\multicolumn{1}{l|}{V4}                    & BGNN \cite{li2021bipartite}                                                  & 75.5                & 37.8                & 41.7                 & 46.9                 \\ \cline{2-6} 
\multicolumn{1}{l|}{}                    & \bf{HL-Net}                 & \bf{78.1} &  \bf{38.9} & \bf{42.2} &\bf{48.1}  \\ \hline
\multicolumn{1}{l|}{\multirow{7}{*}{V6}} 
& MOTIFS \cite{zellers2018neural}                                                  & \multicolumn{1}{c|}{71.6} & \multicolumn{1}{c}{29.9} & \multicolumn{1}{c|}{31.6} & \multicolumn{1}{c}{38.9}  \\
\multicolumn{1}{l|}{}                    & VCTREE \cite{tang2019learning}                                                  & 74.1                & 34.2                & 33.1               & 40.2                 \\
\multicolumn{1}{l|}{}  & RelDN \cite{zhang2019graphical}        & 73.1                & 32.2                & 33.4               & 40.8                 \\    
\multicolumn{1}{l|}{}                    & GPS-Net \cite{lin2020gps}                                                & \multicolumn{1}{c|}{74.8} & \multicolumn{1}{c}{32.9} & \multicolumn{1}{c|}{34.0} & \multicolumn{1}{c}{41.7}  \\
\multicolumn{1}{l|}{}                    & BGNN \cite{li2021bipartite}                                                    & \multicolumn{1}{c|}{75.0} & \multicolumn{1}{c}{33.5} & \multicolumn{1}{c|}{34.2} & \multicolumn{1}{c}{42.1}  \\
\cline{2-6} 
\multicolumn{1}{l|}{}                    & \bf{HL-Net}           & \multicolumn{1}{c|}{\bf 76.5} & \multicolumn{1}{c}{\bf 35.1} & \multicolumn{1}{c|}{\bf 34.7} & \multicolumn{1}{c}{\bf 43.2} \\
\hline
\end{tabular}
\caption{Comparisons with state-of-the-art methods on OI. We adopt the same evaluation metric as in \cite{zhang2019graphical}.}
\label{OI1}
\end{table}

To demonstrate the effectiveness of HL-Net in exploring heterophilic information under occlusion scenarios, we propose to calculate two different R@$K$ metrics for the SGCLS task. More specifically, we decompose the SGCLS task into two subtasks, namely C-SGCLS and S-SGCLS. The former determines the SGCLS performance on triplets where at least one object is heavily occluded by others, \ie, IoU$>$0.5. Otherwise, we term the subtask as S-SGCLS. As shown in Table~\ref{crowd}, HL-Net outperforms all state-of-the-art methods on both tasks. In particular, when compared with MOTIFS \cite{zellers2018neural} using the same ResNeXt-101-FPN backbone, HL-Net achieves an advantage of 5.7$\%$ and 1.8$\%$ at R@100 for C-SGCLS and S-SGCLS, respectively.

\begin{table}
\small
\centering
\setlength{\tabcolsep}{0.8mm}\begin{tabular}{l|ccc|cc|cc}
\hline
\multicolumn{1}{l|}{} &\multicolumn{3}{c|}{Module} & \multicolumn{2}{c|}{SGCLS} & \multicolumn{2}{c}{PREDCLS} \\
\multicolumn{1}{c|}{Exp} &\multicolumn{1}{c}{ART} & \multicolumn{1}{c}{RFP} & \multicolumn{1}{c|}{HMP} & R@50 &  \multicolumn{1}{c|}{R@100}  & \multicolumn{1}{l}{R@50} &   \multicolumn{1}{l}{R@100} \\ \hline\hline
\multicolumn{1}{c|}{1} &\multicolumn{1}{c}{-} & \multicolumn{1}{c}{-} & \multicolumn{1}{c|}{-} &40.1 & \multicolumn{1}{c|}{40.9} &\multicolumn{1}{c}{{65.5}}  &\multicolumn{1}{c}{{67.3}}\\
\multicolumn{1}{c|}{2} &\multicolumn{1}{c}{\checkmark} & \multicolumn{1}{c}{-} & \multicolumn{1}{c|}{-} &41.7 & \multicolumn{1}{c|}{42.5} &\multicolumn{1}{c}{{65.8}} &\multicolumn{1}{c}{{67.6}}  \\
\multicolumn{1}{c|}{3} &\multicolumn{1}{c}{-} & \multicolumn{1}{c}{\checkmark} & \multicolumn{1}{c|}{-} &40.6 & \multicolumn{1}{c|}{41.3} &\multicolumn{1}{c}{{66.4}}&\multicolumn{1}{c}{{68.3}}  \\
\multicolumn{1}{c|}{4} &\multicolumn{1}{c}{-} & \multicolumn{1}{c}{-} & \multicolumn{1}{c|}{\semichecked} &41.2 & \multicolumn{1}{c|}{41.9} &\multicolumn{1}{c}{{65.9}} &\multicolumn{1}{c}{{67.7}}  \\
\multicolumn{1}{c|}{5} &\multicolumn{1}{c}{\checkmark} & \multicolumn{1}{c}{\checkmark} & \multicolumn{1}{c|}{-}&41.8& \multicolumn{1}{c|}{42.7}  &\multicolumn{1}{c}{{66.6}} &\multicolumn{1}{c}{{68.5}}  \\
\multicolumn{1}{c|}{6} &\multicolumn{1}{c}{-} & \multicolumn{1}{c}{\checkmark} & \multicolumn{1}{c|}{\checkmark} &41.3& \multicolumn{1}{c|}{42.1}  &\multicolumn{1}{c}{{66.8}}  &\multicolumn{1}{c}{{68.7}}  \\
\multicolumn{1}{c|}{7} &\multicolumn{1}{c}{\checkmark} &
\multicolumn{1}{c}{-} & \multicolumn{1}{c|}{\semichecked} &42.4& \multicolumn{1}{c|}{43.3} &\multicolumn{1}{c}{{66.1}} &\multicolumn{1}{c}{{67.9}}  \\\hline
\multicolumn{1}{c|}{8} &\multicolumn{1}{c}{\checkmark} & \multicolumn{1}{c}{\checkmark} & \multicolumn{1}{c|}{\checkmark} &\bf{42.6} & \multicolumn{1}{c|}{\bf{43.5}} &\multicolumn{1}{c}{\bf{67.0}}   &\multicolumn{1}{c}{\bf{68.9}}  \\
\hline
\end{tabular}
\caption{Ablation studies. We consistently adopt the same object detection backbone as in \cite{tang2020unbiased}. ``\semichecked" denotes that we only apply HMP to refine the representation of objects.}
\label{ablation1}
\end{table}
Moreover, to demonstrate the robustness of HL-Net to the class imbalance problem on VG, we additionally compare its performance with that of state-of-the-art methods using the Mean Recall metric \cite{chen2019knowledge, tang2019learning}. As shown in Table~\ref{tbam}, HL-Net achieves a notable absolute performance gain without specifically considering the imbalance problem in model design. These results indicate that HL-Net also has advantages to handle the class imbalance problem in SGG. 

\begin{table*}
\centering
\subfloat[Evaluation on the value of $\tau$ in Eq.~\eqref{gamma}.]{
\tablestyle{1.5 pt}{1.3}\begin{tabular}{ccc|clclclcl}
\hline
\multicolumn{3}{c|}{} & \multicolumn{2}{c}{} & \multicolumn{2}{c}{$\tau=0.2$} & \multicolumn{2}{c}{$\tau=0.5$} & \multicolumn{2}{l}{$\tau=0.7$} \\ \hline\hline
 &  &  & \multicolumn{2}{c}{R@20} & \multicolumn{2}{c}{38.5} & \multicolumn{2}{c}{\bf38.8} & \multicolumn{2}{c}{38.6} \\
\multicolumn{3}{c|}{SGCLS} & \multicolumn{2}{c}{R@50} & \multicolumn{2}{c}{42.3} & \multicolumn{2}{c}{\bf42.6} & \multicolumn{2}{c}{42.4} \\
\multicolumn{1}{l}{} &  &  & \multicolumn{2}{c}{R@100} & \multicolumn{2}{c}{43.2} & \multicolumn{2}{c}{\bf43.5} & \multicolumn{2}{c}{43.3} \\ \hline
\end{tabular}}\hspace{5mm}
\subfloat[Evaluation on the number of ART layers $U$.]{
\tablestyle{1.9 pt}{1.3}\begin{tabular}{ccc|clclclclcl}
\hline
\multicolumn{3}{c|}{} & \multicolumn{2}{c}{} & \multicolumn{2}{c}{2-step} & \multicolumn{2}{c}{3-step} & \multicolumn{2}{c}{4-step} & \multicolumn{2}{c}{5-step} \\ \hline\hline
 &  &  & \multicolumn{2}{c}{R@20} & \multicolumn{2}{c}{38.1} & \multicolumn{2}{c}{38.3} & \multicolumn{2}{c}{38.6} & \multicolumn{2}{c}{\bf38.8} \\
\multicolumn{3}{c|}{SGCLS} & \multicolumn{2}{c}{R@50} & \multicolumn{2}{c}{41.9} & \multicolumn{2}{c}{42.1} & \multicolumn{2}{c}{42.3} & \multicolumn{2}{c}{\bf42.6} \\
\multicolumn{1}{l}{} &  &  & \multicolumn{2}{c}{R@100} & \multicolumn{2}{c}{42.8} & \multicolumn{2}{c}{43.0} & \multicolumn{2}{c}{43.3} & \multicolumn{2}{c}{\bf43.5} \\ \hline
\end{tabular}}\hspace{5mm}
\subfloat[Evaluation on the number of RFP layers $K$.]{
\tablestyle{1.5 pt}{1.3}\begin{tabular}{ccc|clclclclcl}
\hline
\multicolumn{3}{c|}{} & \multicolumn{2}{c}{} & \multicolumn{2}{c}{2-step} & \multicolumn{2}{c}{3-step} & \multicolumn{2}{c}{4-step} & \multicolumn{2}{c}{5-step} \\ \hline\hline
 &  &  & \multicolumn{2}{c}{R@20} & \multicolumn{2}{c}{60.3} & \multicolumn{2}{c}{60.5} & \multicolumn{2}{c}{\bf 60.7} & \multicolumn{2}{c}{60.4} \\
\multicolumn{3}{c|}{PREDCLS} & \multicolumn{2}{c}{R@50} & \multicolumn{2}{c}{66.6} & \multicolumn{2}{c}{66.8} & \multicolumn{2}{c}{\bf67.0} & \multicolumn{2}{c}{66.7} \\
\multicolumn{1}{l}{} &  &  & \multicolumn{2}{c}{R@100} & \multicolumn{2}{c}{68.5} & \multicolumn{2}{c}{68.7} & \multicolumn{2}{c}{\bf 68.9} & \multicolumn{2}{c}{68.6} \\ \hline
\end{tabular}}\hspace{5mm}
  \vspace{-2mm}
{\caption{The impact of hyperparameters on the ART and RFP modules, respectively.}\label{ablation_whole}}
\label{tab:2}
\end{table*}

\begin{figure*}[h]
  \begin{center}
    \includegraphics[width=0.95\linewidth]{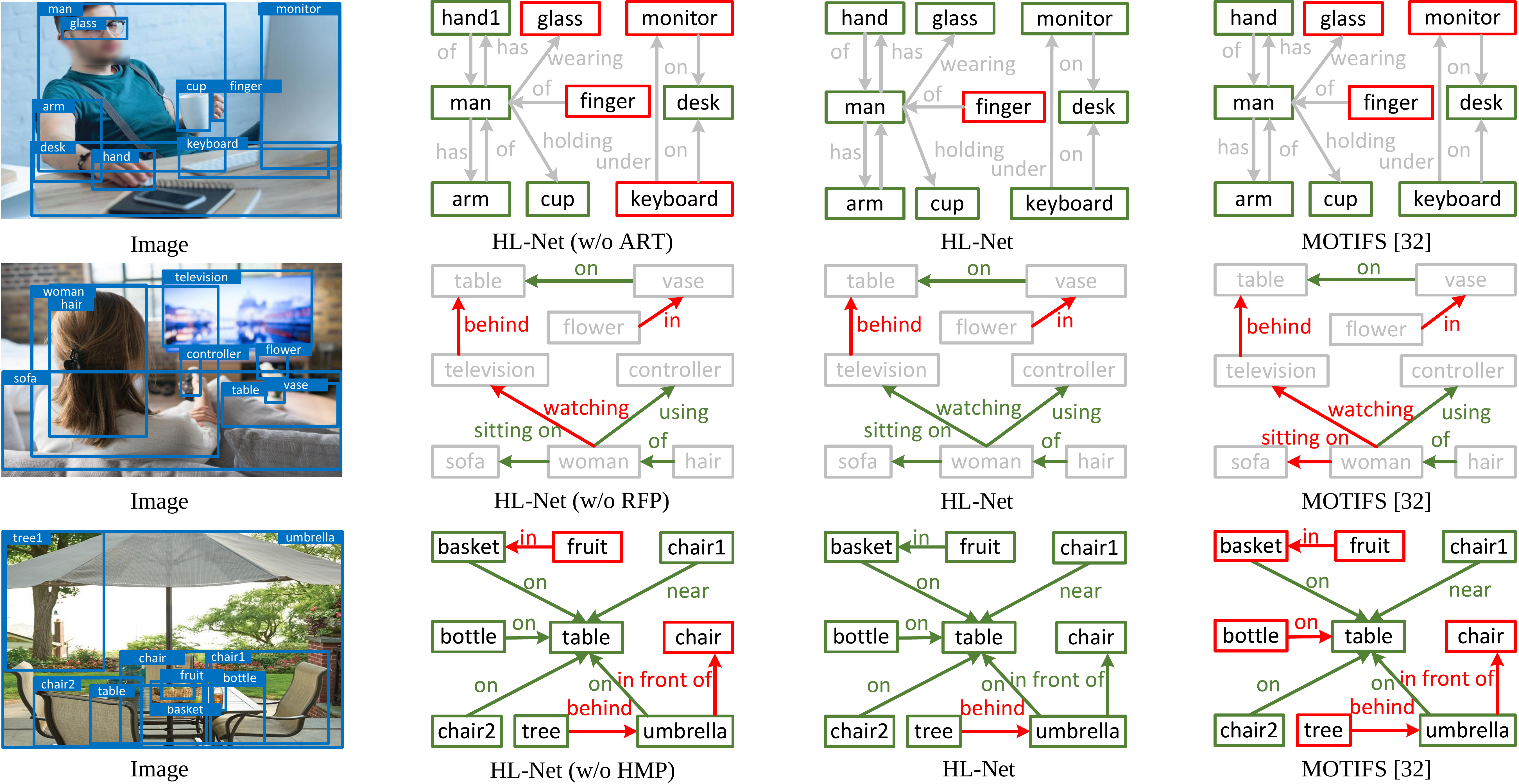}
  \end{center}
  \vspace{-4mm}
\caption{Qualitative comparisons between HL-Net and MOTIFS \cite{zellers2018neural}. Specifically, we show the comparisons at R@100 in the SGCLS setting in the first and third rows. In the second row, we show the comparisons at R@100 in the PREDCLS setting. The green color indicates correctly classified objects or predicates; the red indicates those that have been misclassified. Best viewed in color.}
  \label{qualitative}  
\end{figure*}

\noindent\textbf{Open Images:} 
Table~\ref{OI1} compares the performance of HL-Net with state-of-the-art methods. RelDN is an improved version of the model that won the Google Open Images Visual Relationship Detection Challenge V4. Using the same object detector, HL-Net outperforms RelDN \cite{zhang2019graphical} by 3.5$\%$ and 2.4$\%$ on the overall metric score$_{wtd}$ for OI V4 and V6, respectively. In more detail, on OI V4, HL-Net outperforms RelDN \cite{zhang2019graphical} by 3.2$\%$, 3.4$\%$, and 3.7$\%$ at R@50, wmAP$_{rel}$, and wmAP$_{phr}$, respectively. Moreover, when compared with other approaches that use the same backbone on OI V6, HL-Net consistently achieves the best performance.

\subsection{Ablation Studies}

\noindent\textbf{Effectiveness of the Proposed Modules.} We first perform an ablation study to validate the effectiveness of ART, RFP, and HMP, respectively. The results are summarized in Table \ref{ablation1}. Details of the baseline can be found in Appendix D.2. From Exps 1-8, we can clearly see that the performance improves consistently when more modules are involved. This shows that each module is helpful in promoting the performance of SGG.

\noindent\textbf{Design Choices in ART and RFP.} 
We verify the impact of hyperparameters on the ART and RFP modules. As shown in Table~\ref{ablation_whole}(a), HL-Net achieves the best performance when $\tau$ is set to 0.5 in Eq.~\eqref{gamma}. In Table~\ref{ablation_whole}(b), we compare the performance of HL-Net with different numbers of ART layers, ranging from two to five; it is evident that the performance of HL-Net improves with an increasing number of ART layers (due to limitations on GPU memory size, we only conducted experiments up to five ART layers). In Table~\ref{ablation_whole}(c), we compare the performance of HL-Net with different numbers of RFP layers, ranging from two to five; it is shown that the best performance is achieved when the number of RFP layers is set to four.

\noindent\textbf{Qualitative Evaluation.} Figure~\ref{qualitative} presents a qualitative comparison between HL-Net and MOTIFS \cite{zellers2018neural}. As can be seen from the first row of Figure~\ref{qualitative}, HL-Net makes better predictions than MOTIFS for ``monitor" and ``keyboard" that are hard to recognize from its proposal. Therefore, we owe this performance gain to the ART module that utilizes the heterophilic context to refine the node prediction. In the second row of Figure~\ref{qualitative}, it is shown that HL-Net can identify ``watching" by inferring from their neighboring ones. We give this credit to the RFB module. Finally, in the third row of Figure~\ref{qualitative}, it can be observed that HL-Net has clear advantages in predicting the categories of both nodes and edges under heavy occlusion scenarios (\eg, ``umbrella in front of chair"), via the HMP scheme.

\subsection{Conclusion and Limitations}
Scene graphs are naturally heterophilic. In this paper, we devise HL-Net to comprehensively explore homophily and heterophily for both object and relationship prediction in the SGG. More specifically, the heterophily between nodes is encoded in the message passing via an Adaptive Reweighting Transformer module. The connections between heterophilic relationships are explored by means of a Relationship Feature Propagation module. Moreover, the heterophily and homophily between objects and those between relationships in complicated visual scenes are considered using a Heterophily-aware Message Passing scheme. Extensive experiments on two popular databases justify the effectiveness of HL-Net for SGG. The same as the majority of existing SGG models, one limitation of our method is its dependency on sufficiently labeled data. In the future, we will explore how to train the HL-Net more robustly in the face of a large number of missing annotations.

\noindent\textbf{Broader Impacts.}
SGG can potentially provide valuable assistance for many real-world applications (\eg, autonomous driving). To the best of our knowledge, our work is not harmful in ethical aspects or with future societal consequences.

\noindent{\textbf {Acknowledgment}}. This work is supported by the NSF of China (Nos: 62076101, 62002090), the Program for Guangdong Introducing Innovative and Entrepreneurial Teams (No: 2017ZT07X183).

{\small
\bibliographystyle{ieee_fullname}
\bibliography{egbib}
}

\end{document}